\documentclass[11pt]{article}

\usepackage[preprint]{acl}

\usepackage{times}
\usepackage{latexsym}

\usepackage[T1]{fontenc}
\usepackage[utf8]{inputenc}

\usepackage{microtype}

\usepackage{inconsolata}

\usepackage{graphicx}

\usepackage{booktabs} 
\usepackage{pifont}
\usepackage{threeparttable} % 确保在你的文档导言区加入了这个包
\usepackage{multirow}
\usepackage{subfigure}
\newcommand{\cmark}{\ding{51}} % 对勾
\newcommand{\xmark}{\ding{55}} % 叉号
\usepackage{amsmath}
\usepackage{amssymb}
\usepackage{enumitem}
\usepackage{xcolor}  % 引入颜色宏包
\definecolor{darkgreen}{rgb}{0.0, 0.5, 0.0}

\title{\textit{One Battle After Another}: \\Probing LLMs' Limits on Multi-Turn Instruction Following with a  Benchmark Evolving Framework}

\author{
    Qi Jia\textsuperscript{\rm 1}, 
    Ye Shen\textsuperscript{\rm 1,2},
    Xiujie Song\textsuperscript{\rm 2}, 
    Kaiwei Zhang\textsuperscript{\rm 1}, \\
    \textbf{Shibo Wang}\textsuperscript{\rm 1,3},
    \textbf{Dun Pei}\textsuperscript{\rm 1,2},
    \textbf{Xiangyang Zhu}\textsuperscript{\rm 1}, 
    \textbf{Guangtao Zhai}\textsuperscript{\rm 1,2}\thanks{Corresponding author}\\
    \textsuperscript{\rm 1}Shanghai Artificial Intelligence Laboratory,\\
    \textsuperscript{\rm 2}Shanghai Jiao Tong University, \textsuperscript{\rm 3}Jilin University\\
    \texttt{jiaqi@pjlab.org.cn}
}

\begin{document}
\maketitle
\begin{abstract}

Evaluating LLMs' instruction-following ability in multi-topic dialogues is essential yet challenging. Existing benchmarks are limited to a fixed number of turns, susceptible to saturation and failing to account for users' interactive experience. In this work, we propose a novel framework featuring a three-layer tracking mechanism and a query synthesis agent to mimic sequential user behaviors. Grounded in Flow Theory, we introduce process-centric metrics and terminate a conversational evaluation only upon exhausting user patience. Leveraging this framework, we present EvolIF, an evolving benchmark covering 12 constraint groups. Our analysis reveals deficiencies in failure recovery and fine-grained instruction following, with performance stratification becoming evident as conversational depth increases. GPT-5 demonstrates the most sustained resilience, maintaining a 66.40\% robustness score, outperforming Gemini-3-Pro by 5.59\%, while other models lag behind. Data and code will be released at \url{https://github.com/JiaQiSJTU/EvolIF}.

\end{abstract}

\section{Introduction}
\label{sec:intro}

The rapid advancement of Large Language Models (LLMs) has catalyzed the development of increasingly sophisticated applications, ranging from extended conversational systems~\cite{rakotonirina-etal-2025-tools} to autonomous agent frameworks~\cite{hu2025evaluating}. The efficacy of these systems is fundamentally predicated on an LLM's ability to consistently adhere to instructions throughout conversations spanning multiple topics with evolving constraints. This core capability demands robust long-context processing and stateful memory management. Consequently, designing evaluation frameworks for multi-turn instruction following has emerged as a critical research focus~\cite{he2024multi,kwan2024mt,li-etal-2025-structflowbench}.

\begin{figure}
    \centering
    \includegraphics[width=1.0\linewidth]{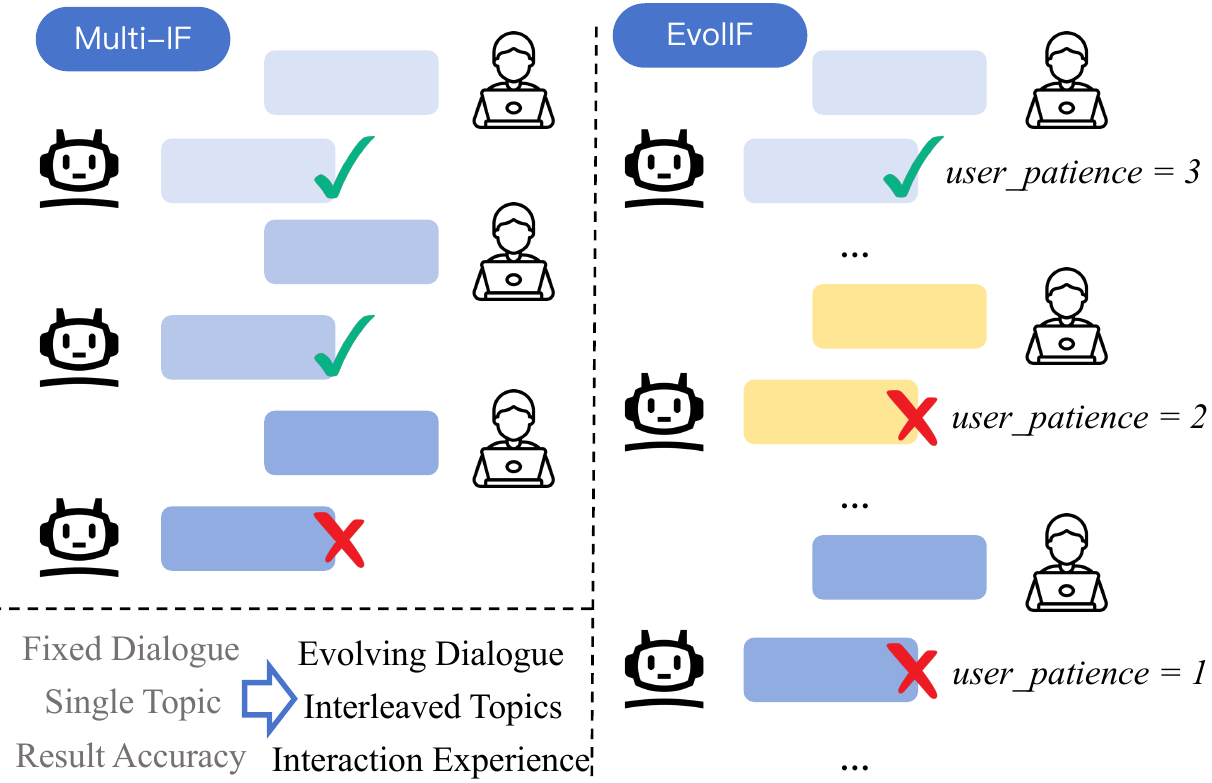}
    \caption{A comparison between Multi-IF and EvolIF. Each color represents a conversational topic. Increasing color saturation signifies the escalating complexity of the instructions as the conversation evolves.} %within the same topic
    \label{fig:illustration}
\end{figure}

\begin{table*}[ht!]
    \centering
    \small

        \begin{tabular}{lcp{1.8cm}p{1.6cm}p{1.8cm}p{1.8cm}}
        \toprule[1pt]
           \textbf{Benchmark}  & \textbf{Avg.\#Turns} & \textbf{Fine-grained Constraint} & \textbf{Multi-Constraint} & \textbf{Topic Transitions} & \textbf{Multi-turn Assessment} \\
        \midrule[1pt]
        IFEval~\cite{zhou2023instruction} & 1 & \cmark & \cmark & \xmark & \xmark \\
        ComplexBench~\cite{wen2024benchmarking} & 1 & \cmark & \cmark & \xmark & \xmark \\
        Multi-IF~\cite{he2024multi} & 3 & \cmark & \cmark & \xmark & \cmark \\
        MT-Eval~\cite{kwan2024mt} & 6.96 & \xmark & \xmark & $\bigcirc$ & \cmark \\
        MT-Bench-101~\cite{bai-etal-2024-mt} & 3.03 & \xmark & \xmark & $\bigcirc$ & \cmark \\
        Meeseeks~\cite{wang2025ask} & 3 & \cmark & \cmark & \xmark & \cmark \\
        EIFBENCH~\cite{zou2025eifbenchextremelycomplexinstruction} & 1 & \cmark & \cmark & \xmark & \xmark \\
        StructFlowBench~\cite{li-etal-2025-structflowbench} & 4.14 & \cmark & \cmark & $\bigcirc$ & \xmark\\
        \midrule[1pt]
        \textbf{EvolIF (ours)}     & $+\infty$ & \cmark & \cmark & \cmark & \cmark \\
        \bottomrule[1pt]
        \end{tabular}
        \caption{Comparisons between EvolIF and other related benchmarks. $\bigcirc$ refers to partially satisfied.
            Avg.\#Turns means the average number of turns in each dialogue sample.
            Fine-grained constraint and multi-constraint denotes the detailed classification of different constraints and whether there exists multiple constraints in a turn.
            Topic transitions indicates whether there are multiple topics discussed in a dialogue.
            Multi-turn assessment checks whether responses to each turn in a dialogue are evaluated.}
        \label{tab:benchmark_comparisons}

\end{table*}

Existing benchmarks suffer from limitations that impede effective evaluation, as exemplified in Fig.~\ref{fig:illustration}.
First, they fail to capture the interaction dynamics~\cite{hao2024meta,zhang2025delrec} and extended duration typical of real-wolrd scenarios. As shown in Table~\ref{tab:benchmark_comparisons}, most benchmarks are restricted to a short interaction window, predominately fewer than 7 turns~\cite{kwan2024mt}, and neglect scenarios involving interleaved topics~\cite{he2024multi,fan2025fairmtbench}.
Second, their static nature leads to rapid performance saturation. As LLMs advance, fixed benchmark challenges are quickly mastered~\cite{he2024multi, bai-etal-2024-mt}. Although some benchmarks offer adjustable complexity~\cite{li2025mtr}, maintaining challenge levels via continuous sample generation incurs prohibitive computational costs for model re-evaluation.
Third, current methodologies overlook the process-centric aspects of user experience. Inheriting the paradigm from single-turn tasks~\cite{zhou2023instruction,zhang-etal-2025-cfbench}, these benchmarks prioritize final-answer accuracy~\cite{he2024multi,li-etal-2025-structflowbench, wang2025ask}. They neglect interaction stability and fail to provide a direct indication of the maximum number of turns that LLMs can maintain high-fidelity instruction following.

To overcome these shortcomings, we propose a novel and extensible framework for the dynamic generation and process-centric evaluation of complex multi-turn dialogues. 
Our approach decouples user queries into underlying intentions and surface form.
Intention is tracked via a three-layer mechanism that simulates dynamic user behaviors, while the surface form is synthesized by an agent equipped with an LLM-based generator and rigorous validity checkers.
Besides, we move beyond single-turn accuracy to emphasize process-centric experience, drawing upon the Flow Theory~\cite{csikszentmihalyi1990flow}. We introduce the notion of \textit{patience} to model user stickiness to a conversation, where consecutive frustrations lead to a dialogue termination. And we define a suite of process-centric metrics to quantify user experience such as endurance and robustness.

Leveraging this framework, we introduce EvolIF, a benchmark grounded on 541 topics, 12 groups of commonly-adopted constraint groups and 500 diverse user styles. Through an evaluation of 10 leading LLMs, we observe a distinct performance stratification. GPT-5 and Gemini-3-Pro establish a commanding lead, whose process-centric scores are nearly double or triple of open-source models. Besides, LLMs share a common and steepest performance drop at around turn 5 and 12, revealing critical bottlenecks in their ability to manage accumulated constraints and complex state transitions.

To sum up, the contributions of this paper are:

\begin{itemize}
    \item We propose an extensible framework for dynamically generating multi-turn evaluation datasets that resist saturation.
    \item We introduce EvolIF to assess the limits of LLMs' long-context management and instruction-following abilities.
    \item We analyze state-of-the-art LLMs, offering insights into their robustness in prolonged dialogues and identifying critical limitations to guide future optimization.
\end{itemize}

\section{Related Work}

\subsection{Multi-turn Dialogue Benchmarks}

Existing work for benchmarking LLMs in multi-turn dialogues can be categorized as follows:

First, {script-based evaluations}~\cite{li-etal-2025-structflowbench,deshpande-etal-2025-multichallenge,jia2025simulbench} utilize static conversational logs, derived either from human-bot interactions or simulated histories, to assess a model's response to the final user query. While this approach ensures controlled and consistent LLM comparison, it fails to capture the interactive nature of dialogue, where a model's prior responses fundamentally influence the conversational trajectory.

Second, a line of work employs {pre-defined templates}~\cite{zheng2023judging,fan2025fairmtbench,han2025can}. This approach is labor-intensive, requiring significant human effort to design fixed user query sequences. Consequently, these benchmarks face scalability limitations regarding conversational depth and are susceptible to saturation as models become overly optimized to the test set over time.

Third, researchers have explored using LLMs as user simulators~\cite{zhu2024reliable,sekulic-etal-2024-reliable} and evaluation methods based on conversations between LLMs~\cite{duan-etal-2024-botchat,zhao-etal-2025-auto}. Nevertheless, such interactions are prone to uncontrolled divergence and exhibit inherent biases, such as family bias~\cite{wataoka2025selfpreferencebiasllmasajudge}.

In contrast, our framework integrates the structural rigor of pre-defined evaluations with the linguistic richness of LLM-based synthesizers, enabling a dynamic generation of theoretically unlimited dialogue turns.

\subsection{Instruction Following Benchmarks}

Research on instruction following is primarily divided into single-turn and multi-turn paradigms.

One line of work assesses models' capabilities within increasingly intricate single-turn interaction. Early benchmarks like CIF~\cite{li-etal-2024-cif} evaluated a single constraint per instruction. Subsequent work has evolved to incorporate multiple constraints~\cite{zhou2023instruction, jiang-etal-2024-followbench, wen2024benchmarking,he2024can} or multiple tasks~\cite{chen-etal-2024-sifo, zou2025eifbenchextremelycomplexinstruction}.

A parallel stream of work benchmarks models' instruction adherence across multiple turns. Multi-IF~\cite{he2024multi} extends IFEval to 3 turns, while MultiTurnInstruct~\cite{han2025can} employs pre-defined templates for diverse scenarios. StructFlowBench~\cite{li-etal-2025-structflowbench} leverages 6 structure types to curate complex dialogue histories. Other studies focus on specialized abilities, such as self-correction~\cite{wang2025ask}, or domain-specific tasks like code generation~\cite{wang2025codeif}.

Our framework offers a more flexible and scalable data synthesis process for mitigating saturation issues inherent in static benchmarks. Moreover, by integrating a suite of process-oriented metrics, we offer a more holistic, multi-faceted performance analysis that prioritizes the user's experience.

\section{A Benchmark Evolving Framework}

\begin{figure*}[ht!]
    \centering
    \includegraphics[width=0.9\linewidth]{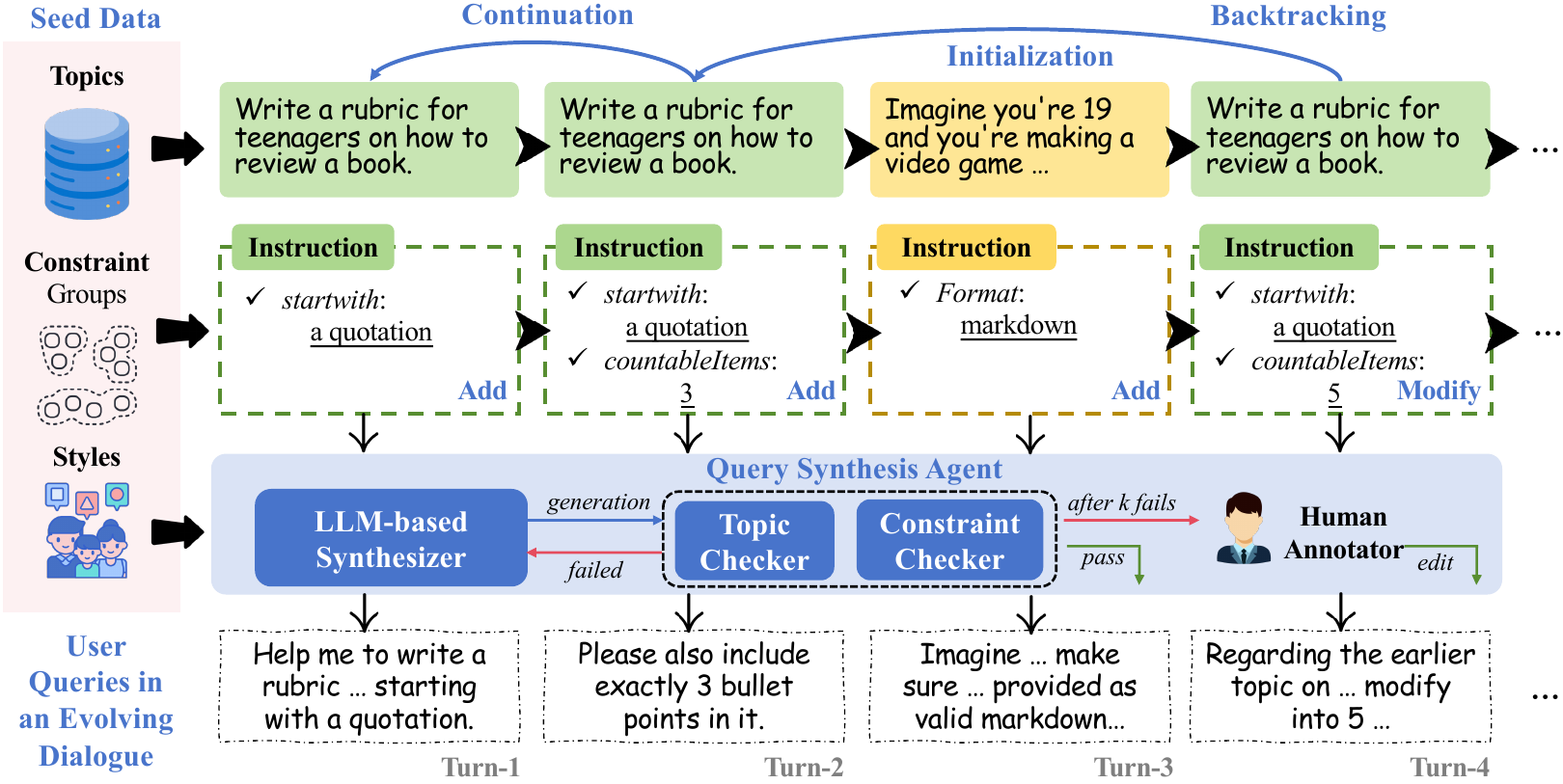}
    \caption{Overview of the Benchmark Evolving Framework.}
    \label{fig:overview}
\end{figure*}

\subsection{Overview}

We propose a novel framework comprises three integral components: a dynamic data synthesis engine, an adaptive evaluation protocol, and a suite of process-centric metrics. Crucially, the framework can be flexibly adapted to diverse domains by simply preparing seed topics and defining in-domain constraints. An overview of the proposed architecture is illustrated in Figure~\ref{fig:overview}.

The \textbf{dynamic data synthesis engine} is designed to generate consecutive user queries by orchestrating a \textbf{three-layer tracking mechanism} and \textbf{a query synthesis agent}. Based on the intuition of decomposing user queries, this mechanism manages topics, instructions, and constraints separately. It enables a flexible simulation of user behaviors, such as instruction refinement, topic switching and backtracking. 
The query synthesis agent transforms simulated state and a sampled user style into a final query. Query validity is ensured by an iterative verification loop involving an LLM-based synthesizer and different checkers, with human oversight as the ultimate gatekeeper. This dynamic composition allows for the generation of a theoretically infinite stream of queries.

Flow theory~\cite{csikszentmihalyi1990flow} points out that users enter a psychological state of immersion which can withstand minor disruptions but collapses under prolonged failure. 
Underpinned by this theory, we employ an \textbf{adaptive evaluation protocol} where the length of a dialogue is contingent on model performance, governed by a ``patience'' threshold. High-performing models face progressively longer and more challenging threads, while repeated failures will deplete patience and trigger termination. In this way, our benchmark remains a persistent challenge for advanced models, resisting from saturation.

Furthermore, we broaden the evaluation scope from the final-answer accuracy to a holistic conversational experience via \textbf{process-centric metrics}. \textit{Endurance} quantifies the sustainable conversation depth, \textit{recovery} measures the model's resilience in realigning with user intent after a mistake, and \textit{robustness} evaluates the stability of instruction adherence across turns.

\subsection{Data Synthesis Engine}

We model a user query $q_t$ at turn $t$ as a tuple $(\mathcal{U}_t, s_t)$. The structured intention $\mathcal{U}_t$ is precisely managed by the {three-layer tracking mechanism}, providing an unambiguous ground truth that ensures the validity of the evaluation. On top of it, the surface form $s_t$ is stochastically generated to capture linguistic diversity with a query synthesis agent, mimicking real users with various styles.

\subsubsection{Three-layer Tracking Mechanism}

To dynamically manage the evolution of the dialogue state and simulate the full spectrum of evolving user intentions, we further decompose $\mathcal{U}_t$ into a hierarchy of three interconnected components: \textbf{Topics}, \textbf{Instructions}, and \textbf{Constraints}. Each layer serves a different level of semantic control, collectively forming the foundation of our framework.

\textbf{Topic Layer}\quad A topic $T \in \mathbb{T}$ represents a subject or event under discussion. 
It captures the conversational flow, particularly in longer interactions involving topic switching and interleaved sub-dialogues~\cite{li2025revisiting}. Our framework maintains a history of active topics $H_T=(T_1, T_2, ...)$.

\textbf{Instruction Layer}\quad Each topic $T$ is associated with an instruction state $\mathcal{I}_T$, which encapsulates a set of atomic constraints $\{c_1, c_2, \dots, c_k\}$. Throughout a dialogue, $\mathcal{I}_T$ evolves via the addition, deletion, or modification of its constituent constraints, simulating how a user's goal shifts over time.

\textbf{Constraint Layer}\quad Constraints $\mathbb{C}$ are categorized into $m$ mutually exclusive groups. In other words, a group $G_i$ contains constraints that cannot be simultaneously satisfied, defined with the satisfaction set $\mathcal{S}(c)$:
\begin{equation*}
    \forall c_a, c_b \in G_i \text{ with } c_a \neq c_b, \quad \mathcal{S}(c_a) \cap \mathcal{S}(c_b) = \emptyset
\end{equation*}
Consequently, $\mathcal{I}_T$ is restricted to contain at most one constraint from any given group $G_i$, to avoid creating unachievable requirements:
\begin{equation*}
    |\mathcal{I}_T \cap G_i| \le 1, \quad \forall i \in \{1, \dots, m\}
\end{equation*}

A conversation script is constructed turn-by-turn through a stochastic process. At each turn $t$, the state transitions from $S_{t-1}$ to $S_t$ via three steps:

\textbf{Topic Selection}\quad The topic for $T_t$ is determined by the transition function $\phi_T$ operating on $H_T$: either \textit{continue} the current topic ($T_t = T_{t-1}$), introduce a \textit{new} topic ($T_t \notin H_T$), or \textit{backtrack} to a historical topic ($T_t \in H_T$).

\textbf{Instruction Evolution}\quad Once $T_t$ is selected, its associated instruction $\mathcal{I}'_{t}$ undergoes structural evolution. $\phi_{\mathcal{I}}$ updates the set of constraints through addition, modification or removal. 

\textbf{Constraint Evolution}\quad Parameters of individual constraints are randomly altered by $\phi_c$, yielding the final instruction for the the current turn:  % within $\mathcal{I}'_{T_t}$
    \begin{equation}
        \mathcal{I}_t = \phi_c(\phi_{I}(\mathcal{I}'_{t})).
    \end{equation}

\subsubsection{Query Synthesis Agent}

The generated script, represented by a sequence of topic-instruction pairs $\{(T_t, \mathcal{I}_t)\}_{t=1}^N$, is rendered into natural utterances by the Query Synthesis Agent. 
It consists of \textbf{an LLM-based synthesizer} and a series of \textbf{checkers} to ensure output validity.

To bolster linguistic diversity and stylistic consistency, a persona style $\delta$ is randomly specified for each dialogue. We utilize adaptive prompting strategies to generate contextually coherent queries at turn $t$ with a piecewise function as follows:

\begin{equation}
p_t = 
\begin{cases} 
    f_{\text{new}}(T_t, \mathcal{I}_t, \delta), & \text{if } T_t \text{ is new,} \\
    f_{\text{continue}}(\mathcal{I}_t, \mathcal{I}_{t-1},\delta), & \text{if } T_t = T_{t-1}, \\
    f_{\text{backtrack}}(T_t, \mathcal{I}_t, \mathcal{I}_{t-1},\delta), & \text{otherwise.}
\end{cases}
\end{equation}
$f_{\text{new}}$ introduces a new topic with its initial instructions. $f_{\text{continue}}$ highlights modifications to existing requirements. $f_{\text{backtrack}}$ signals a reversion to a prior topic while introducing updated instructions.

Topic checkers and constraint checkers are incorporated to ensure the accurate convey of the user's intent. The query will be re-generated unless it passes all of them for maximum $k$ iterations. Otherwise, it is flagged for human review. 

In summary, this synthesis process yields an infinite stream of extensible dialogues, providing a foundation for fair and reproducible multi-turn instruction-following evaluation.

\subsection{Evaluation Protocol}

Our evaluation protocol is adaptive and designed to mirror real-world user interactions, premised on Flow Theory~\cite{csikszentmihalyi1990flow} and the cooperative principles of dialogue~\cite{grice1975logic}. 
Repeated failures by a conversation partner serve as a primary catalyst for user frustration, leading to the eventual disengagement~\cite{ang2002prosody,hernandez-caralt-etal-2025-stupid}.

To address this, we first support dynamical adjustment of the session length. Dialogues in the constructed benchmark can be extended as long as the model follows instructions successfully. %This aligns with the observation that users are willing to engage in prolonged conversations with competent partners.

Furthermore, we introduce a patience score $P$, initialized to a maximum value $P_{max}$, to simulate user tolerance. Our protocol dictates that the dialogue terminates after a sequence of consecutive failures. Specifically, after each turn $t$, $P$ is updated based on the model's performance.
\begin{equation}
P_t = 
\begin{cases} 
    P_{t-1} - 1, & \text{if failed,} \\
    P_{max}, & \text{otherwise.}
\end{cases}
\end{equation}
The evaluation session concludes when the patience score is exhausted, i.e., $P_t = 0$.

\subsection{Evaluation Metrics}

Conventional metrics, such as \textbf{Constraint Satisfaction Rate (CSR)} and \textbf{Instruction Satisfaction Rate (ISR)}\cite{zhang-etal-2025-cfbench, li-etal-2025-structflowbench}, focus primarily on outcome accuracy. To capture the nuances of the conversational process, we introduce a suite of process-centric metrics. Given a benchmark of $D$ dialogues, these metrics are defined below (see Appendix\ref{app:basic_metrics} for details).

\textbf{Endurance (EDR)} measures conversational longevity under varying degrees of strictness. Let $N_d$ be the total number of turns in dialogue $d$. % the number of conversation turns under different strictness. 
\begin{itemize}[leftmargin=0.5cm]
    \item \textbf{Length (EDR$_{\text{len}}$):} The number of turns a model sustains before termination, regardless of their correctness. This measures pure persistence.
    \begin{equation*}
        \text{EDR}_{\text{len}} = \frac{1}{D} \sum_{d=1}^{D} N_d
    \end{equation*}
    \item \textbf{Accuracy (EDR$_{\text{acc}}$):} The cumulative constraint satisfaction rate accumulated over the conversation, rewarding partial correctness. %provides the most granular view of endurance, which
    \begin{equation*}
        \text{EDR}_{\text{acc}} = \frac{1}{D} \sum_{d=1}^{D} \sum_{t=1}^{N_d} \frac{|C_{d,t}^{\text{sat}}|}{|\mathcal{I}_{d,t}|}
    \end{equation*}
    \item \textbf{Success (EDR$_{\text{succ}}$):} The number of turns where the model perfectly satisfies all instruction. %This measures productive endurance.
    \begin{equation*}
        \text{EDR}_{\text{succ}} = \frac{1}{D} \sum_{d=1}^{D} \sum_{t=1}^{N_d} \mathbb{I}(|\mathcal{I}_{d,t}| = |C_{d,t}^{\text{sat}}|)
    \end{equation*}

    \item \textbf{Longest Satisfaction Sequence (EDR$_{\rm lss}$):} The maximum number of \textit{consecutive} turns in which instructions are perfectly satisfied. %of perfect instruction satisfaction.
        \begin{equation*}
        \begin{aligned}
            & \text{EDR}_{\text{lss}}  = \frac{1}{D} \sum_{d=1}^{D} \max_{1 \le j \le k \le N_d} \\
            & \left\{ k-j+1 \mid \forall t: j \le t \le k, |\mathcal{I}_{d,t}| = |C_{d,t}^{\text{sat}}| \right\}
        \end{aligned}
        \end{equation*}
        
\end{itemize}

\textbf{Recovery (REC)} assesses a model's resilience by measuring its ability to succeed after one or more failures within the patience $P$. 
\begin{equation}
\begin{aligned}
        &\text{REC} =\\
        &\frac{1}{D}\sum_{d=1}^{D} \frac{\sum_{t=2}^{N_d} \mathbb{I}(\text{ISR}_{d,t-1}=0 \land \text{ISR}_{d,t}=1)}{\sum_{t=2}^{N_d} \mathbb{I}(\text{ISR}_{d,t-1}=0)} 
\end{aligned}
\end{equation}

\textbf{Robustness (ROB)} measures the overall reliability of a model, defined as the macro-average of the ISR across all dialogues.
\begin{equation*}
    \text{ROB} = \frac{1}{D} \sum_{d=1}^{D} \left( \frac{1}{N_d} \sum_{t=1}^{N_d} \mathbb{I}(|\mathcal{I}_{d,t}| = |C_{d,t}^{\text{sat}}|) \right)
\end{equation*}

\begin{table*}[ht]
    \centering
    \small
    \begin{tabular}{l|cccc|ccc|c}
    \toprule
       \textbf{Models}  & \textbf{EDR\textsubscript{len}} & \textbf{EDR\textsubscript{acc}} & \textbf{EDR\textsubscript{succ}} & \textbf{EDR$_{lss}$} & \textbf{CSR (\%)} & \textbf{ISR (\%)} & \textbf{REC (\%)} & \textbf{ROB (\%)}\\
    \midrule
    GPT-5 & \textbf{19.32} & \textbf{17.11} & \textbf{14.09} & \textbf{8.80}  & \textbf{88.57} & \textbf{72.91}  & \textbf{29.09} & \textbf{66.40} \\
    Gemini-3-Pro & \underline{16.36} & \underline{14.11} & \underline{11.41} & \underline{7.17} & \underline{86.22} & \underline{69.72}  & \underline{27.50} & \underline{60.81} \\
    MiniMax-M2 & 11.75 & 9.19 & 7.17 & 4.77 & 78.22 & 60.98 & 24.29  & 54.54\\
    Kimi-K2  & 10.16 & 7.82 & 5.99 & 4.13 & 76.92 & 58.99 & 19.52 & 48.43\\
    Qwen3-235B & 10.02 & 7.66 & 5.80 & 3.97 & 76.43 & 57.88 & 21.15 & 47.47\\
    Grok-4-Fast  & 9.52 & 7.29 & 5.50 & 4.13 & 76.58 & 57.77 & 16.01 & 46.03\\
    DeepSeek-V3.2 & 8.64 & 6.32 & 4.62 & 3.38 & 73.15 & 53.47 & 15.87 & 44.42\\
    Seed-1.6 & 8.21 & 5.78 & 4.20 & 2.95 & 70.44 & 51.18 & 15.90 & 39.43\\
    Llama-4-Maverick & 8.10 & 5.21 & 3.90 & 2.76 & 64.37 & 48.15 & 19.05 & 39.37 \\
    Mistral-Large-3 & 7.86 & 5.37 & 3.91  & 2.83 & 68.34 & 49.70  & 15.79 & 38.56 \\

    \bottomrule
    \end{tabular}
    \caption{Main results on the EvolIF benchmark. Higher is better for all metrics. Best results are bolded and the second best results are underlined.}
    \label{tab:main_results}
\end{table*}

\section{Experimental Setup}

\subsection{EvolIF Benchmark}

Leveraging our framework, we introduce \textbf{EvolIF}, a benchmark for assessing multi-turn instruction-following capability of LLMs.

We first curated its core assets: topics, constraints and styles. We collected 541 dialogue topics from IFEval~\cite{zhou2023instruction}, manually removing the attached constraints to isolate the core task scenarios and subjects. To support our dynamic generation process, we assigned a set of customized keywords for each topic. Concurrently, we consolidated constraints from prior works~\cite{zhou2023instruction,li-etal-2025-structflowbench} and our own construction, and systematically re-categorized them into 12 mutually exclusive groups based on semantic intention. These contain 9 objective constraints assessed by rules and 3 subjective constraints measured with an LLM judge. 
Moreover, we gathered 500 styles by prompting GPT-4.1 with personas from~\citet{nvidia2025}. More in Appendix~\ref{app:seed_data}.

We guarantee the quality of the benchmark through the following considerations. 
To ensure integrity and complexity, we applied an automated filter to discard trivial samples, removing dialogues where the average number of constraints over the first 20 turns was less than two. 
To mitigate family bias~\cite{spiliopoulou2025play} introduced by a single synthesizer, we adopted GPT-4.1, Gemini-2.5-Flash and DeepSeek-V3.1 as synthesizers to generate dialogue sessions with $k=3$ trials.

The final EvolIF benchmark contains 150 distinct dialogues. Unlike traditional static benchmarks that rely on a large number of short, finite-turn samples, EvolIF prioritizes conversational depth and endurance. It's extensible nature, combined with a rich variety of dynamic behaviors, including instruction evolution, topic switching, and backtracking, makes it a challenging and future-proof testbed for evaluating the long-term capabilities of advanced models. The default patience score was set to $P_{max}=3$.

\subsection{Evaluated Models}

We conducted evaluation on ten state-of-the-art large language models from different institutions. They include GPT-5-2025-08-07, Gemini-3-Pro-Exp~\cite{comanici2025gemini}, DeepSeek-V3.2-Exp~\cite{liu2024deepseek}, Kimi-K2-Instruct-0905~\cite{team2025kimi}, Qwen-235B-A22B-Instruct-2507~\cite{yang2025qwen3}, Grok-4-Fast-Reasoning, Llama-4-Maverick, Seed-1.6-Thinking-250715, MiniMax-M2 and Mistral-Large-3. All of the models were evaluated with corresponding default settings. Code and data will be released. 

% @online{openai2025gpt5,
%   author   = {{OpenAI}},
%   title    = {Introducing GPT-5},
%   year     = {2025},
%   url      = {https://openai.com/index/introducing-gpt-5/},
% }
% @online{openai2025gpt41,
%   author   = {{OpenAI}},
%   title    = {Introducing GPT-4.1 in the API
% },
%   year     = {2025},
%   url      = {https://openai.com/index/gpt-4-1/},
% }
% @online{x2025grok4,
%   author   = {{X.ai}},
%   title    = {Grok 4
% },
%   year     = {2025},
%   url      = {https://x.ai/news/grok-4},
% }
% @online{meta2025llama,
%   author   = {{Meta}},
%   title    = {The Llama 4 herd: The beginning of a new era of natively multimodal AI innovation},
%   year     = {2025},
%   url      = {https://ai.meta.com/blog/llama-4-multimodal-intelligence/},
% }
% @online{bytedance2025seed,
%   author   = {{ByteDance}},
%   title    = {Seed1.6 Tech Introduction},
%   year     = {2025},
%   url      = {https://seed.bytedance.com/en/seed1\_6},
% }
% @online{mistral2025,
%   author   = {{Mistral AI}},
%   title    = {Medium is the new large},
%   year     = {2025},
%   url      = {https://mistral.ai/news/mistral-medium-3},
% }

\section{Results and Analysis}

This section first presents the main results using our multi-faceted metrics, followed by an analysis of conversational endurance and a fine-grained breakdown by constraint groups. We also examine the impact of user patience on perceived capability and evaluate ranking stability across sample sizes. More analyses of system prompts, synthesis models, and user styles are in the appendices.

\begin{figure*}[h!]
    \centering
    \includegraphics[width=1.0\linewidth]{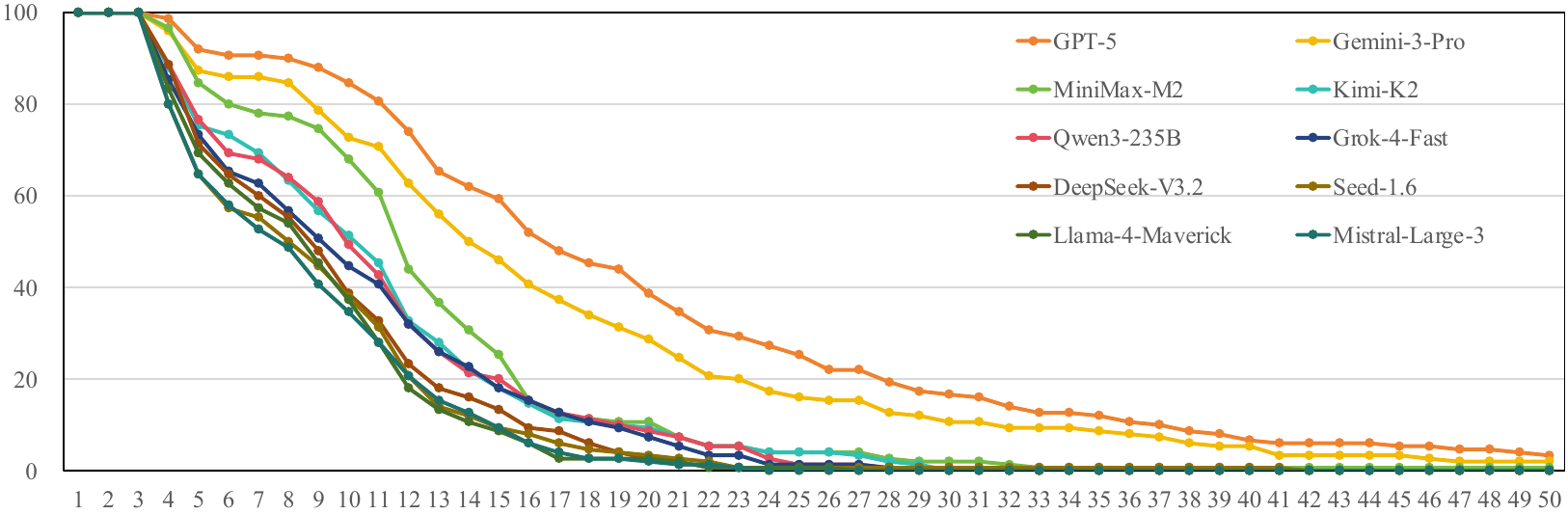}
    \caption{Dialogue survival curves for all ten evaluated models. The y-axis shows the percentage of initial sessions still active at each turn. Slower decay rates indicate higher conversational endurance and resilience.}
    \label{fig:survival_curve}
\end{figure*}

\subsection{Main Results}

The performance of LLMs on EvolIF is presented in Table~\ref{tab:main_results}. Our analysis reveals a distinct stratification in multi-turn instruction-following capabilities. GPT-5 establishes itself as the state-of-the-art, with Gemini-3-Pro following closely behind. These two models demonstrate a superior performance, achieving process-centric scores that are double or even triple of subsequent models. MiniMax-M2 emerges as the most competitive open-source LLMs, forming a second tier alongside Kimi-K2, Qwen3-235B and Grok-4-Fast. The rest constitute the third tier, indicating substantial difficulties in maintaining long and accurate conversations.

\textbf{Multi-Turn Capability and Endurance}
The EDR metrics provide a quantitative measure of the models' upper limits for sustained instruction following. The disparity in EDR\textsubscript{succ}, which focuses on the productive responses, is pronounced. GPT-5 sustains an average of 14.09 fully successful turns, whereas this figure drops to approximately 6 turns for mid-tier models and merely 3.90 turns for the weakest model, Llama-4-Maverick. Furthermore, EDR$_{lss}$, which focuses on uninterrupted performance, poses a higher bar for capability. GPT-5 exhibits exceptional stability with a correct streak of 8.80 turns, far surpassing the leading open-source model, MiniMax-M2, with 4.01 turns.

\textbf{Accuracy and Resilience}
Regarding instruction accuracy, CSR and ISR metrics reinforce the observed performance hierarchy. In terms of resilience, REC scores are universally lower than 30\%, with the top-performing GPT-5 achieving only 29.09\%. Grok-4 struggles the most on this aspect among the second-tier models, while Llama-4-Maverick demonstrates strong recovery capability despite its overall weaker ranking. This widespread lack of resilience is a primary factor leading to premature dialogue termination, limiting models' practical usability in long conversations.

\textbf{Overall Robustness}
ROB serves as a holistic indicator of reliability, effectively distinguishing model capabilities while other metrics misght show ambiguity. For instance, while Qwen-3-235B and Grok-4-Fast exhibits similar performance on CSR and ISR, Grok-4-Fast suffers from weaker recovery capabilities. This deficiency is captured by ROB, which reveals a performance gap of 1.44\% between the two models, highlighting ROB's value as a comprehensive evaluative score.

\subsection{Dialogue Survival Analysis}

To visualize and compare the long-term memory management capabilities of the models over time, we tracked the percentage of active dialogue sessions remaining at each turn, up to a maximum of 50 turns. This yields a dialogue survival curve for each model, as depicted in Figure~\ref{fig:survival_curve}.
A performance stratification between the top-3 models and the rest emerges at turn 4, right after the fast exhaustion of the user patience. Initially, MiniMax-M2 demonstrates instruction-following capabilities comparable to GPT-5 and Gemini-3-Pro. However, its performance drops dramatically after 10 turns, becoming indistinguishable from second-tier models by turn 15.

The survival curve confirms that the primary differentiator between model tiers is not merely single-turn accuracy, but resilience to accumulating complexity. Turns 4-5 and 11-12, where models exhibit their common and steepest drops, serve as practical indicators of a shared complexity ceiling. At these points, the LLMs' ability to track interleaved topics and instructions begins to collapse. Notably, top-tier models lose 50\% of their dialogue sessions around turn 15, whereas other models consistently hit this wall around the 10th turn, highlighting a critical area for future improvement.

\begin{figure*}
    \centering
    \includegraphics[width=\linewidth]{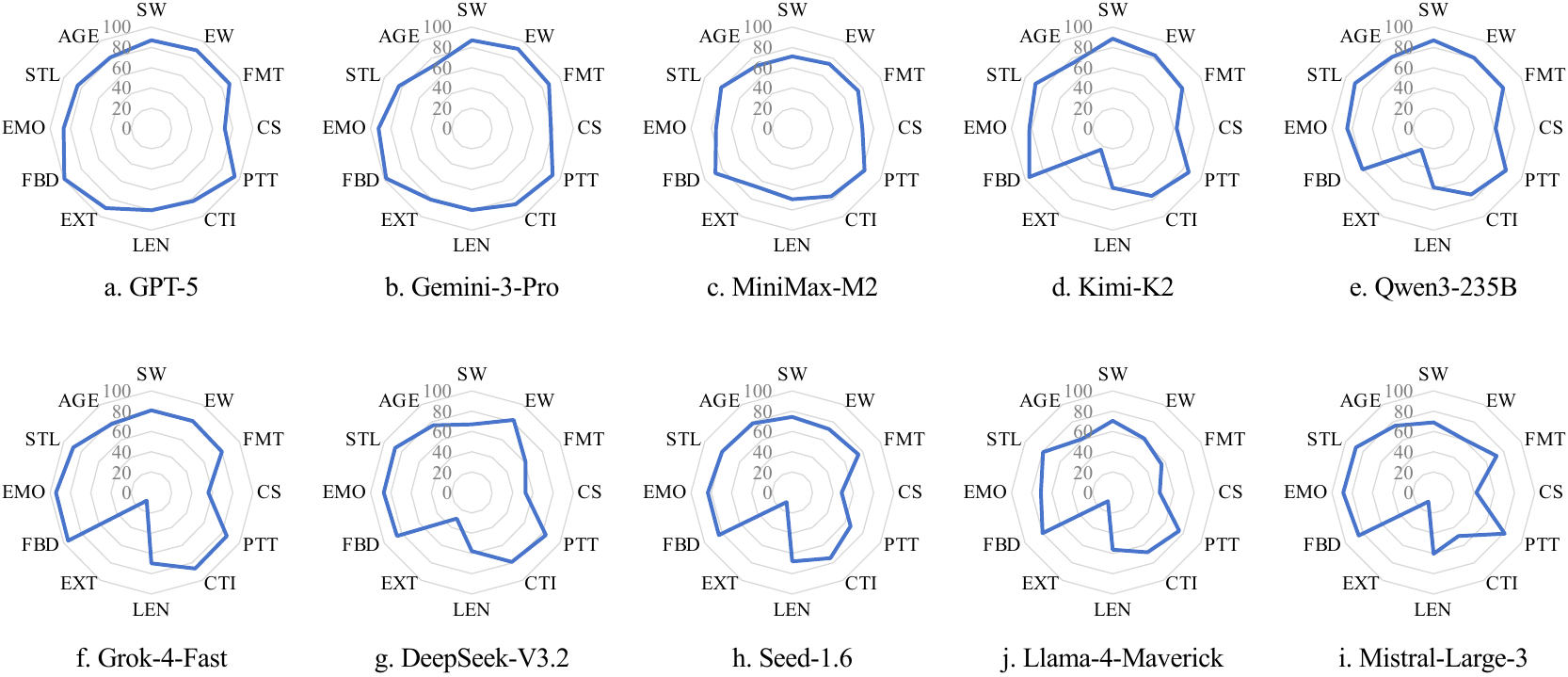}
    \caption{Instruction Satisfaction Rate (\%) per Constraint Group on the EvolIF benchmark.}
    \label{fig:per_group_accuracy}
\end{figure*}

\subsection{Fine-Grained Analysis of Constraints}
\label{sec:fine-grained-constraint}

We provide a detailed breakdown of model performance across the 12 constraint groups in EvolIF to identify shared difficulties and reveal model-specific weaknesses. Detailed statistics are in Appendix~\ref{app:constraint_group}, and the unique performance profiles of different models are depicted in Figure~\ref{fig:per_group_accuracy}.

Among objective constraints, LLMs perform best on FBD and PTT. These constraints are essentially binary checks, requiring the model to simply include or exclude specific content. Conversely, the most challenging groups are EXT and CS, which reveal a significant gap between the best and weakest models. These constraints demand global planning and state tracking at the word and character levels throughout a response. Subjective constraints also present challenges. While LLMs are adept at handling different emotions and styles, they struggle to adapt to the preferences of different age groups.

GPT-5 and Gemini-3-Pro demonstrate strong, well-rounded performance, topping the rankings on most objective constraints. However, they lag behind Qwen3-235B and Grok-4-Fast on subjective tasks. MiniMax-M2 does not achieve outstanding performance in any single category but ultimately outperforms the remaining models that exhibit spiky profiles.

\subsection{Analysis on the User's Patience}
\label{sec:ablation_patience}

\begin{table}[t]
    \centering
    \small
    \begin{tabular}{l|ccc}
    \toprule
    \textbf{Models} & 1 & 2 & 3 \\
    \midrule
    GPT-5 & 7.03 & 11.19 & 17.11\\
    Gemini-3-Pro & 5.40 & 9.16 & 14.11\\
    MiniMax-M2 & 3.76 & 6.66 & 9.19 \\
    Kimi-K2 & 3.46 & 5.52 & 7.82 \\
    Qwen3-235B & 3.17 & 5.36 & 7.66 \\
    Grok-4-Fast & 3.53 & 5.20 & 7.29 \\
    DeepSeek-V3.2 & 2.82 & 4.50 & 6.32 \\
    Seed-1.6 & 2.60 & 4.20 & 5.78 \\
    Llama-4-Maverick & 2.02 & 3.65 & 5.21\\
    Mistral-Large-3 & 2.33 & 3.91 & 5.45 \\
    
    \bottomrule
    \end{tabular}
    \caption{The effect of the patience score ($P$) on EDR$_{\text{acc}}$.}
    \label{tab:patience_ablation}
\end{table}

Table~\ref{tab:patience_ablation} illustrates the impact of user tolerance on conversational endurance. By varying the patience score $P$, we simulate a spectrum of user temperaments to assess model robustness in sustaining long-term interactions. %this diversity in user tolerance, the analysis reveals how robustly each model can maintain a positive user experience throughout a long conversation.
Raising the patience threshold from 1 to 3 roughly doubles the average dialogue length across all models. Crucially, this relaxation amplifies performance gaps. The lead of GPT-5 over Llama-4-Maverick expands from 5.01 to 11.90 turns. This trend indicates that models with strong self-correction abilities, i.e.,  high REC, disproportionately benefit from the added buffer provided by increased patience.

\subsection{Sensitivity to Sample Sizes}

\begin{table}[ht]
    \centering
    \small
    % \begin{threeparttable}
    \begin{tabular}{p{2.05cm}|p{0.6cm}p{0.6cm}p{0.6cm}p{0.6cm}p{0.6cm}}
    \toprule
       \textbf{Models }  & 30 & 60 & 90 & 120 & 150 \\
    \midrule
    GPT-5 & 71.21 & 68.29 & 68.60 & 68.42 & 66.40\\
    Gemini-3-Pro & 62.04 & 63.02 & 61.88 & 61.36 & 60.81 \\
    MiniMax-M2 & 55.37 & 57.34 & 56.03 & 55.80 & 54.54 \\
    Kimi-K2 & 46.61\textcolor{red}{$\downarrow$1} & 47.24\textcolor{red}{$\downarrow$1} & 48.38\textcolor{red}{$\downarrow$1} & 48.60\textcolor{red}{$\downarrow$1} & 48.43 \\
    Qwen3-235B & 49.83\textcolor{darkgreen}{$\uparrow$1} & 48.49\textcolor{darkgreen}{$\uparrow$1} & 49.56\textcolor{darkgreen}{$\uparrow$1} & 48.61\textcolor{darkgreen}{$\uparrow$1} & 47.47 \\
    Grok-4-Fast & 45.04 & 46.34 & 46.68 & 45.86 & 46.03 \\
    DeepSeek-V3.2 & 40.47\textcolor{red}{$\downarrow$1} & 42.72 & 44.06 & 45.16 & 44.42 \\
    Seed-1.6 & 42.05\textcolor{darkgreen}{$\uparrow$1} & 41.08 & 38.44\textcolor{red}{$\downarrow$1} & 38.67\textcolor{red}{$\downarrow$2} & 39.43 \\
    Llama-4 & 36.41 & 37.92\textcolor{red}{$\downarrow$1} & 38.44 & 39.14 & 39.37 \\
    Mistral-Large-3 & 35.44 & 38.62\textcolor{darkgreen}{$\uparrow$1} & 40.16\textcolor{darkgreen}{$\uparrow$2} & 40.30\textcolor{darkgreen}{$\uparrow$2} & 38.56 \\
    \hline
    PLCC & 98.08 & 99.22 & 99.55 & 99.69 & -\\
    \bottomrule
    \end{tabular}
    \caption{Ranking stability with different number of samples according to ROB(\%). Arrows indicate relative ranking shifts compared to the full dataset, and PLCC calculates the corresponding Pearson Correlation (\%).}
    \label{tab:ranking_stability}
\end{table}

Unlike previous works that rely on a large volume of test samples, we prioritize extending interaction length to differentiate LLM capabilities. This raises the question of whether the 150 samples in EvolIF are sufficient to yield a stable LLM ranking. To address this, we compare LLM rankings across varying sample sizes in Table~\ref{tab:ranking_stability}. The results reveal that rankings only fluctuate locally among similar models, while the overall hierarchy stabilizes with as few as 30 samples.

\section{Conclusion}

In this work, we introduced an extensive framework for multi-turn instruction following that integrates dynamic data synthesis, an adaptive evaluation protocol, and a suite of process-oriented metrics. Built upon this framework, our benchmark, EvolIF, moves beyond static evaluations to measure the crucial dimensions of conversational experience. Our experiments reveal a clear performance hierarchy among leading LLMs, uncovering a universal weakness in error recovery and a systemic struggle with fine-grained constraints requiring planning during the generation.

\section*{Limitations}

Our framework aims to simulate authentic user behaviors to probe the boundaries of LLMs in real-world scenarios. Currently, we primarily target textual instruction following, merging rigorous verifiability with linguistic diversity. EvolIF encompasses both objective and subjective constraints. Moving forward, we intend to incorporate multi-modality, tool usage, and personalization of topics and instructions to facilitate a more comprehensive evaluation of LLMs and MLLMs.

Besides, following previous work, such as Arena-Hard~\cite{li2024crowdsourced} and MT-Bench~\cite{zheng2023judging}, we choose the LLM-as-a-judge approach for subjective constraint evaluation. We acknowledge the inherent limitations of this approach, such as family bias~\cite{spiliopoulou2025play}. In this work, we utilize it as a widely-adopted verifier and prompt it with detailed instructions. Notably, we observed no significant family bias using GPT-4.1, given that it did not disproportionately prefer GPT-5 across subjective tasks. This judge could also be replaced by targeted classifiers. Developing more robust verification methods lies beyond the scope of this paper.


% \section*{Limitations}

% \section*{Acknowledgments}



\bibliography{custom}

\newpage
\appendix
\section{Metrics}
\label{app:basic_metrics}
\subsection{Basic Metrics}

Following prior work \cite{zhang-etal-2025-cfbench, li-etal-2025-structflowbench}, we quantify overall instruction-following accuracy. Let $N$ be the total number of turns replied by the model. We adopt the following metrics:

\textbf{Constraint Satisfaction Rate (CSR)} measures the average satisfaction rate of individual constraints across all $N$ turns. It provides a fine-grained assessment of how well the model adheres to specific requirements.
\begin{equation*}
    \text{CSR} = \frac{1}{N} \sum_{t=1}^{N} \frac{|C_t^{\text{sat}}|}{|\mathcal{I}_t|}
\end{equation*}
where $C_t^{\text{sat}} \subseteq \mathcal{I}_t$ is the set of constraints satisfied by the model's output at turn $t$.

\textbf{Instruction Satisfaction Rate (ISR)} offers a strictly turn-level perspective compared to CSR. It calculates the proportion of turns in which the model successfully satisfies \textit{all} constraints, measuring overall reliability of a model on a turn-by-turn basis:
\begin{equation*}
    \text{ISR} = \frac{1}{N} \sum_{t=1}^{N} \mathbb{I}(|\mathcal{I}_t| = |C_t^{\text{sat}}|)
\end{equation*}
where $\mathbb{I}(\cdot)$ is the indicator function.

\subsection{Process-Centric Metrics}

\begin{figure}[h!]
    \centering
    \includegraphics[width=0.92\linewidth]{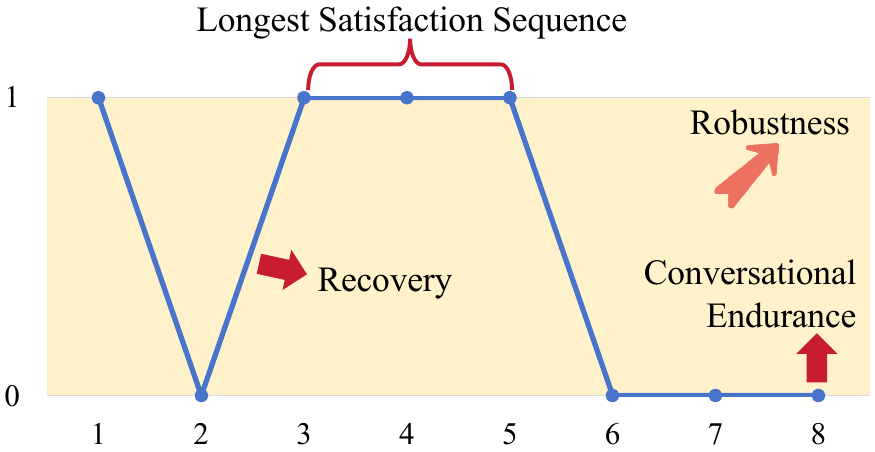}
    \caption{Process-centric Metrics.}
    \label{fig:metrics}
\end{figure}

We propose a suite of evaluation metrics, providing a holistic and complementary view of a model's conversational competence. \textbf{EDR}  quantifies various dimensions of conversational longevity, while \textbf{REC} captures the critical capability of self-correction following errors. \textbf{ROB} offers a unified score for overall reliability. An illustration of these process-based metrics is presented in Figure~\ref{fig:metrics}. The ranges of these metrics are explained as follows.

EDR theoretically ranges from a minimum of $P_{max}$ to infinity. Since a conversation persists as long as the patience score permits, the minimum possible length corresponds to the initial patience threshold $P_{max}$, occurring in the case of immediate consecutive failures, with no upper limit for a perfectly performing model.

The REC metric falls within the range of $[0, 1)$. A model that consistently fails to recover from errors will rapidly exhaust its patience and terminate the dialogue, naturally driving its REC score toward 0.

The ROB metric is bounded within the range $[0, 1)$. While typically bounded between 0 and 1, the practical upper bound in our framework is constrained by the patience mechanism. Since every session must eventually terminate with $P_{max}$ consecutive failures, a model cannot achieve a perfect ROB of 1 in a finite session. Specifically, for a dialogue of length $N$, the maximum attainable ROB is $\frac{N-P_{max}}{N}$. However, for an infinitely capable model, this upper bound converges to 1 as the conversation length $N$ approaches infinity.

\section{Seed Data Preparation}
\label{app:seed_data}

\subsection{Topic}

We collected 541 prompts from IFEval~\cite{zhou2023instruction}. One annotator was tasked with removing the instructions and constraints from each prompt to extract the corresponding dialogue topic. Customized keywords for each topic were then generated by GPT-4.1. Subsequently, each topic and its keywords were verified by two additional annotators. Modifications were iteratively adopted until the data was accepted by both of them.

\begin{table*}[h!]
\centering
\small
\begin{tabular}{lp{5.6cm}p{6cm}}
\toprule
\textbf{Group Name} & \textbf{Constituent Constraints} & \textbf{Description} \\
\midrule
\textit{Objective constraints:}\\
StartWith (SW) & Letter, Emoji, Keyword, Quotation & Controls what the response must begin with. \\
EndWith (EW)& Letter, Emoji, Keyword, Quotation & Controls what the response must end with. \\
Format (FMT) & JSON, HTML, XML, CSV, Markdown & Requires the response to adhere to a specific structured format. \\
Case (CS) & All Uppercase, All Lowercase, Min Uppercase Ratio & Enforces rules on the capitalization of letters in the response.\\
Punctuation (PTT) & MustInclude, MustNotInclude & Governs the inclusion or exclusion of specific punctuation marks. \\
CountableItems (CTI) & Bullet Points & Requires an exact number of bullet points in the response. \\
Length (LEN) & Word Count, Paragraph Count, Character Count, Sentence Count & Controls the length of the response based on various units.\\
Existence (EXT) & MustContain (with exact counts) & Requires specific keywords to appear an exact number of times. \\
Forbidden (FBD) & MustNotContain & Forbids the inclusion of specific keywords. \\
\midrule
\textit{Subjective constraints:}\\
Emotion (EMO) & Happy, Sad, Neutral, Angry, Excited, Frustrated & Sets the emotional tone. \\
Style (STL) & Formal, Informal, Active Voice, Passive Voice & Defines the writing style.\\
Age (AGE) & Child, Youth, Adult, Senior & Tailors to the target age group.\\
\bottomrule
\end{tabular}
\caption{The constraint groups in the EvolIF benchmark.}
\label{tab:constraint_groups}
\end{table*}

\subsection{Constraint Group}
\label{app:constraint_group}

We collected constraints from existing works~\cite{zhou2023instruction,li-etal-2025-structflowbench} and related research. Ultimately, the constraints were classified into 12 groups as shown in Table~\ref{tab:constraint_groups}. 9 of them are objective, verifiable with rule-based functions using existing parser packages or regular expressions. The remaining 3 subjective groups draw inspiration from prior work on style transfer~\cite{heylighen1999formality}, emotion recognition~\cite{busso2008iemocap} and age bias analysis~\cite{liu2024generation}. These subjective constraints are measured by adopting GPT-4.1 as a judge, following previous work~\cite{li2024crowdsourced,zheng2023judging}. Specifically, GPT-4.1 is prompted to score constraint satisfaction on a scale of 1 to 10 with detailed explanations. We consider scores greater than 6 as accepted.

\subsection{Style}

% more details
We randomly selected 500 personas from \citet{nvidia2025}. Then, we employed GPT-4.1 to infer the most plausible language style and tone each persona would use in daily conversation with 3 to 5 descriptive phrases. These phrases serve as inputs to the Query Synthesis Agent to facilitate the generation of diverse and engaging user queries.

\section{Data Quality Analysis}
% checkers 
% Scanned files: 150
% Total turns: 4519

% instruction_success:
%   true: 4462, false: 57, false_and_user_query_changed: 23
% topic_success:
%   true: 4486, false: 33, false_and_user_query_changed: 10

EvolIF comprises 150 dialogues and currently supports 4519 turns. Only 1.26\% of synthesized queries failed to pass the Constraint Checkers. Among them, 40.36\% are adjusted by human annotators, while the remainder were identified as false negative warnings stemming from linguistic diversity not covered by the checkers. Besides, 0.73\% of queries triggered the Topic Checker with 30.30\% being modified. These statistics reflect the reliability of the synthesized queries by LLMs, particularly when reinforced by human annotators as the final safeguard.

\section{Performance on Different Constraints}

\begin{table*}[ht!]
    \centering
    \scriptsize
    \begin{tabular}{l|ccccccccc|ccc}
    \toprule
      \textbf{Models}  & \textbf{SW} & \textbf{EW} & \textbf{FMT} & \textbf{CS} & \textbf{PTT} & \textbf{CTI} & \textbf{LEN} & \textbf{EXT} & \textbf{FBD} & \textbf{EMO} & \textbf{STL} & \textbf{AGE} \\
    \midrule
    GPT-5 & \underline{86.99} & \underline{89.20} & \textbf{88.55} & \underline{72.08} & \textbf{94.73} & 82.52 & \textbf{80.42} & \textbf{90.67} & \textbf{99.20} & 86.69 & 84.50 & \underline{81.13} \\
    Gemini-3-Pro & 86.96 & \textbf{90.74} & \underline{87.50} & \textbf{77.71} & \underline{91.81} & \underline{86.27} & \underline{80.19} & \underline{80.60} & \underline{97.86} & \underline{92.13} & 83.45 & 72.21  \\
    MiniMax-M2 & 71.12 & 73.31 & 74.80 & 69.18 & 82.15 & 77.10 & 69.59 & 66.92 & 87.65 & 75.38 & 81.28 & 71.30 \\
    Kimi-K2 & \textbf{88.56} & 83.06 & 79.04 & 62.66 & 86.44 & 76.60 & 58.44 & 23.79 & 95.25 & 82.39 & 87.92 & 75.68 \\
    Qwen3-235B & 86.94 & 80.50 & 79.72 & 61.34 & 82.76 & 75.00 & 57.82 & 23.83 & 80.50 & 84.98 & \textbf{89.20} & \textbf{82.28} \\
    Grok-4-Fast & 80.93 & 81.34 & 80.49 & 55.80 & 84.75 & \textbf{86.52} & 69.97 & 9.27 & 94.86 & \textbf{94.22} & \underline{89.05} & 77.77 \\
    DeepSeek-V3.2 & 67.15 & 82.49 & 60.75 & 52.71 & 84.18 & 78.98 & 57.94 & 29.58 & 85.33 & 86.99 & 87.72 & 76.49  \\
    Seed-1.6 & 74.40 & 71.93 &75.23 & 48.40 & 66.49 & 74.87 & 67.91 & 11.17 & 83.64 & 83.41 & 79.91 & 78.31\\
    Llama-4 & 70.59 & 61.61 & 55.28 & 46.29 & 75.54 & 68.00 & 56.13 & 9.66 & 79.90 & 70.97 & 79.30 & 60.87 \\
    Mistral-Large-3 & 69.01 & 60.38 & 71.79 & 42.47 & 81.25 & 49.42 & 60.39 & 10.19 & 84.82 & 88.89 & 88.34 & 75.82\\
    \midrule
    Average & 78.27 & 77.46 & 75.32 & 58.86 & 83.01 & 75.53 & 65.88 & 35.57 & 88.90 & 84.61 & 85.07 & 75.19\\
    \bottomrule
    \end{tabular}
    \caption{Instruction Satisfaction Rate (\%) per Constraint Group on the EvolIF benchmark. Best results are in bold and the second best results are underlined.}
    \label{tab:per_group_accuracy}
\end{table*}

Table~\ref{tab:per_group_accuracy} provides a detailed breakdown of model performance across 12 pre-defined constraint groups. This fine-grained analysis is crucial for diagnosing the primary obstacles LLMs face in multi-turn instruction following, allowing us to both identify the inherent difficulty of different constraint types and reveal model-specific weaknesses. We classify the objective ones into three categories.

\textbf{Easiest Constraints:} Models perform best on FBD and PTT. These constraints are essentially binary checks, requiring the model to simply include or exclude specific content. The high accuracy indicates that models possess robust capacity for such straightforward instructions. Similarly, SW and EW constraints also show high performance with over 77\% accuracy across models, as they emphasize local control at the text's boundaries and do not necessitate global planning over the entire generation process.

\textbf{Moderate Constraints:} FMT and CTI fall into a middle tier of difficulty. These two constraints share the similarity on assessing the model's ability to generate structured output, which is a critical skill for applications like code generation and agent-based systems. While models can often produce the correct general structure, they frequently struggle with syntactic precision, especially when these constraints are combined with others in a dialogue. 

\textbf{Hardest Constraints:} The most challenging group by a significant margin is EXT, where a large performance gap separates GPT-5 and Gemini-3-Pro from all other models. This highlights that while models can be prompted to include keywords, they are exceptionally poor at adhering to specific frequency counts. Following closely in difficulty are LEN and CS. These constraints all demand a form of global planning and state tracking over the fine-grained words and characters throughout generation. This suggests that while models are fluent producers of text, their ability to maintain adherence to fine-grained structural and quantitative rules remains a significant limitation.

Regarding subjective constraints, which focus on overall linguistic expression, EMO and STL fall into the easiest group, whereas AGE proves more challenging. Alternatively, since we adopted GPT-4.1 for assessing these subjective aspects, this result may also reflect that LLMs show lower agreement on age-related features compared to emotion and style. More targeted analysis of this phenomenon will be considered in future work.

\section{The Role of the System Prompt}

\begin{table*}[ht!]
    \centering
    \small
    \begin{tabular}{l|cc|cccc|cc}
    \toprule
       \textbf{Models \& Conditions}  & \textbf{CSR (\%)} & \textbf{ISR (\%)} & \textbf{EDR\textsubscript{len}} & \textbf{EDR\textsubscript{acc}} & \textbf{EDR\textsubscript{succ}} & \textbf{LSS} & \textbf{ROB (\%)} & \textbf{REC (\%)} \\
    \midrule
    \textbf{Gemini-3-Pro} & \textbf{85.72} & \textbf{68.16} & \textbf{15.64} & \textbf{10.66} & \textbf{13.41} & \textbf{6.64} & \textbf{61.49} & \textbf{24.98} \\
    \quad w.o. system prompt & 78.04 & 62.04 & 10.96 & 8.55 & 6.8 & 5.02 & 52.22 & 21.52 \\
    \midrule
    \textbf{DeepSeek-V3.2} & \textbf{73.48} & \textbf{55.26} & \textbf{8.94} & \textbf{6.57} & \textbf{4.94} & \textbf{3.76} & \textbf{46.27} & 16.25 \\
    \quad w.o. system prompt & 70.81 & 53.43 & 8.16 & 5.78 & 4.36 & 3.10 & 43.27 & \textbf{17.68} \\
    \midrule
    \textbf{Llama-4} & 62.15 & 46.54 & 7.22 & \textbf{4.49} & 3.36 & 2.50 & 35.65 & 15.87 \\
    \quad w.o. system prompt & \textbf{65.48} & \textbf{48.70} & \textbf{7.72} & 3.76 & \textbf{5.06} & \textbf{2.74} & \textbf{37.54} & \textbf{17.41} \\
    \bottomrule
    \end{tabular}
    \caption{A comparison of results with or without using a system prompt.}
    \label{tab:system_prompt}
    
\end{table*}

Our evaluation methodology utilizes a system prompt that explicitly outlines the task requirements for the model. To assess its impact, we randomly select 50 samples and compare the default setting with a "w.o. system prompt" condition in Table~\ref{tab:system_prompt}. The results indicate that a high-level system prompt provides essential guidance that anchors the model's behavior and improves instruction adherence, particularly for more capable models. The top-tier model, Gemini-3-Pro, suffers the most substantial decline without a system prompt. EDR\textsubscript{len} of it drops by nearly 5 turns, and ROB falls by over 9.27\%. DeepSeek-V3.2 shows a more modest reduction of 3\% in ROB. In contrast, Llama-4 suffers appears to be hindered by the additional system prompt, exhibiting an improvement of approximately 2\% in ROB when the prompt is removed.

\section{Sensitivity to Different LLM Synthesizers}

\begin{table*}[ht]
    \centering
    \small
    \begin{tabular}{l|cc|cccc|cc}
    \toprule
       \textbf{Models \& Conditions}  & \textbf{CSR (\%)} & \textbf{ISR (\%)} & \textbf{EDR\textsubscript{len}} & \textbf{EDR\textsubscript{acc}} & \textbf{EDR\textsubscript{succ}} & \textbf{LSS} & \textbf{ROB (\%)} & \textbf{REC (\%)} \\
    \midrule
    \textbf{GPT-5}$^{*}$ & \underline{88.57} & \underline{72.91} & {19.32} & \underline{17.11} & \underline{14.09} & {8.80} & {66.40} & \underline{29.09} \\
    \quad w. GPT-4.1 &  88.41 & 72.81 & 18.98 & 16.78 & 13.82 & \underline{9.14} & \textbf{67.27} & \textbf{31.28}\\
    \quad w. Gemini-2.5-Flash & 88.26 & 70.66 & \underline{19.36} & 17.09 & 13.68 & 7.90 & 64.85 & 28.60 \\
    \quad w. DeepSeek-V3.1 & \textbf{89.02} & \textbf{75.23} & \textbf{19.62} & \textbf{17.47} & \textbf{14.76} & \textbf{9.36} & \underline{67.07} & 27.39 \\
    \midrule
    \textbf{Gemini-3-Pro}$^{*}$ & {86.22} & \underline{69.72} & \underline{16.36} & \underline{14.11} & \underline{11.41} & {7.17} & {60.81} & {27.50} \\
    \quad w. GPT-4.1 & \underline{87.02} & \textbf{70.63} & \textbf{17.98} & \textbf{15.65} & \textbf{12.70} & \textbf{7.70} & \underline{62.63} & \underline{27.79} \\
    \quad w. Gemini-2.5-Flash & \textbf{87.06} & 69.04 & 15.88 & 13.83 & 11.02 & \underline{7.18} & \textbf{62.85} & \textbf{29.65}\\
    \quad w. DeepSeek-V3.1 & 84.40 & 68.99 & 15.22 & 12.85 & 10.5 & 6.62 & 56.94 & 25.06 \\    
    \midrule
    \textbf{DeepSeek-V3.2}$^{*}$ & 73.15 & 53.47 & 8.64 & 6.32 & 4.62 & 3.38 & 44.42 & \underline{15.87} \\
    \quad w. GPT-4.1 & \underline{73.37} & \underline{53.67} & \underline{8.98} & \underline{6.59} & \underline{4.82} & \underline{3.60} & \underline{47.26} & \textbf{18.07}  \\
    \quad w. Gemini-2.5-Flash & 70.73 & 48.87 & 7.94 & 5.62 & 3.88 & 2.64 & 38.09 & 14.44 \\
    \quad w. DeepSeek-V3.1 & \textbf{75.07} & \textbf{57.33} & \textbf{9.00} & \textbf{6.76} & \textbf{5.16} & \textbf{3.90} & \textbf{47.91} & 15.10 \\
    \bottomrule
    \end{tabular}
    \caption{Ablation studies on the impact of the instruction synthesis model. All experiments were run with a patience score of $P=3$.}
    \label{tab:synthesizer_ablation}
    
\end{table*}

We analyze the performance of models on different subsets of samples generated by different synthesizers in Table~\ref{tab:synthesizer_ablation}. Note that these results are simultaneously influenced by the user styles randomly sampled for each instance. Nevertheless, we observe that Gemini-3-Pro achieves better performance on data synthesized by Gemini-2.5-Flash, whereas other models find it more challenging.

\begin{figure}
    \centering
    \includegraphics[width=0.8\linewidth]{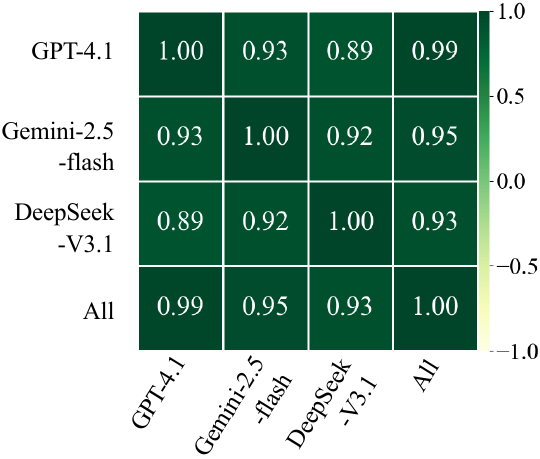}
    \caption{Spearman correlation of LLM ROB scores across samples generated by different synthesizers.} %Spearman Correlation of LLMs' ROB scores with samples generated by different synthesizers.}
    \label{fig:correlation}
\end{figure}

We further calculate the correlation of ROB scores across the ten LLMs between different subsets and the full test dataset. The results in Figure~\ref{fig:correlation} reflect strong positive correlations in model performance across various synthesizers. In summary, the model ranking on EvolIF proves to be robust, particularly when considering the mixed synthesis strategy employed in our benchmark.

\section{Performance trend with different styles}

\begin{figure}[t!]
    \centering
    \includegraphics[width=\linewidth]{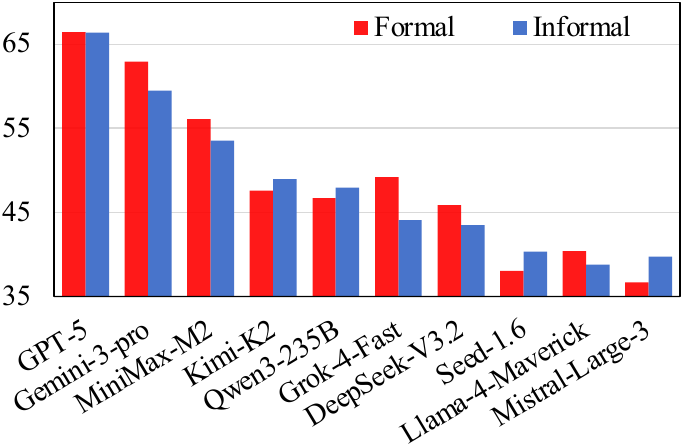}
    \caption{ROB (\%) scores across different user style categories.}
    \label{fig:persona_ablation}
\end{figure}

We prompted GPT-4.1 to classify the dataset into two categories, i.e., formal and informal, based on the linguistic style of user queries. The performance comparison is presented in Figure 7. GPT-5 remains relatively stable across different styles, whereas other LLMs exhibit varying performances. Gemini-3-Pro, MiniMax-M2, Grok-5-Fast, DeepSeek-V3.2, and Llama-4 favor a more formal linguistic style characterized by clear intentions. Conversely, the remaining models show a preference for informal styles, where user queries are typically more engaging.

% \section{Example Appendix}
% \label{sec:appendix}

% This is an appendix.

\end{document}